\documentclass[5pt]{article}

\usepackage[letterpaper]{geometry}
\usepackage{spconf,amsmath,epsfig}
\usepackage{enumitem}

\usepackage[british]{babel}
\usepackage{graphicx}
\usepackage{amssymb}
\usepackage{algpseudocode}
\usepackage{booktabs}
\usepackage{url}
\usepackage{array}
\usepackage{balance}
\usepackage{multirow}
\usepackage{caption}
\usepackage{algorithm}
\usepackage{sidecap}
\usepackage{mathrsfs}

\usepackage{fancyhdr}
\newcolumntype{P}[1]{>{\centering\arraybackslash}p{#1}}
\begin{document}\sloppy

\title{Evolving Boxes for Fast Vehicle Detection}
\name{Li Wang$^{1,2}$, Yao Lu$^3$, Hong Wang$^2$, Yingbin Zheng$^2$, Hao Ye$^{2*}$\thanks{$^*$Corresponding author.}, Xiangyang Xue$^1$\thanks{This work was supported in part by grants from NSFC (\#61602459, \#61572138, and \#U1611461) and STCSM's program (\#16511104802, \#16JC1420401, and \#17YF1427100).}}
\address{$^1$Shanghai Key Lab of Intelligent Information Processing, School of Computer Science, \\Fudan University, Shanghai, China\\
$^2$Shanghai Advanced Research Institute, Chinese Academy of Sciences, Shanghai, China\\
$^3$School of Computer Science and Engineering, University of Washington, Seattle, USA\\
wangli16@fudan.edu.cn, luyao@cs.washington.edu,\\ \{wang\_hong,zhengyb,yeh\}@sari.ac.cn, xyxue@fudan.edu.cn}
\maketitle

\begin{abstract}
We perform fast vehicle detection from traffic surveillance cameras. A novel deep learning framework, namely Evolving Boxes, is developed that proposes and refines the object boxes under different feature representations. Specifically, our framework is embedded with a light-weight proposal network to generate initial anchor boxes as well as to early discard unlikely regions; a fine-turning network produces detailed features for these candidate boxes. We show intriguingly that by applying different feature fusion techniques, the initial boxes can be refined for both localization and recognition. We evaluate our network on the recent DETRAC benchmark and obtain a significant improvement over the state-of-the-art Faster RCNN by 9.5\% mAP. Further, our network achieves 9-13 FPS detection speed on a moderate commercial GPU.
\end{abstract}
\begin{keywords}
Vehicle detection, deep neural networks
\end{keywords}

\section{Introduction}
\label{sec:intro}

\begin{figure}[t]
  \centering
  \includegraphics[width=0.49\linewidth]{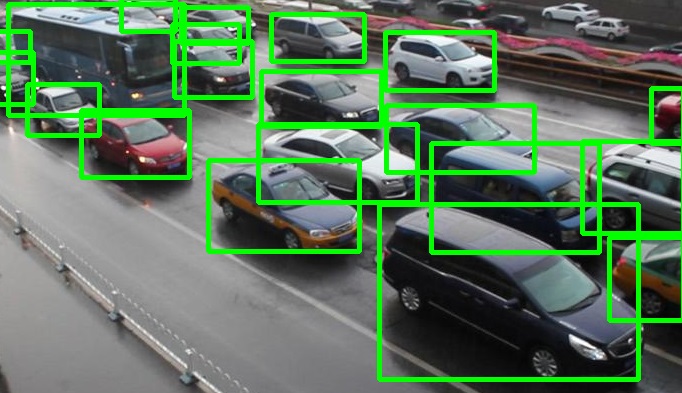}
  \includegraphics[width=0.48\linewidth]{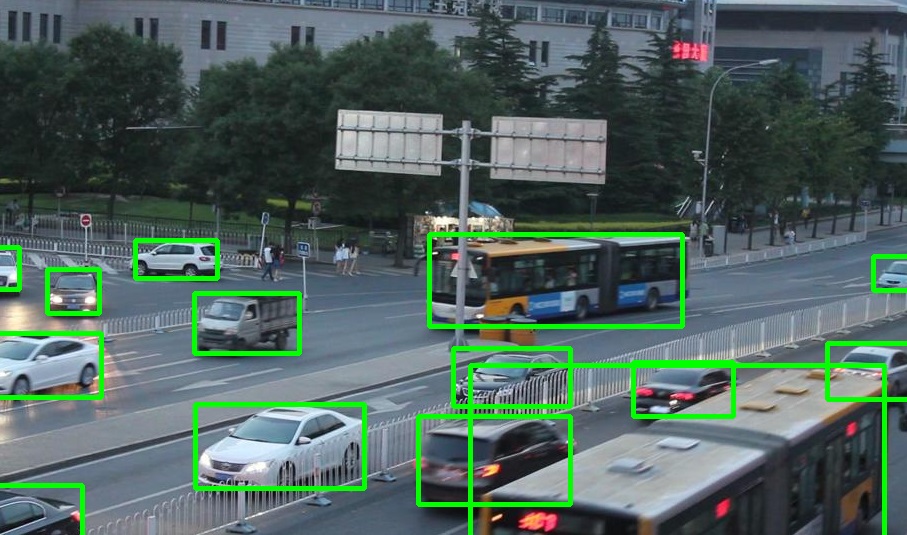}
  \caption{Vehicle detection in real traffic surveillance cameras is challenging. There is often a large variation in occlusion and lightning conditions as well as vehicle pose and scale patterns. We demonstrate two examples from the DETRAC vehicle detection dataset \cite{DETRAC:CoRR:WenDCLCQLYL15}; heavy traffic, different weather conditions and different vehicle types are shown in these examples.  }
  \label{fig:gt_face}
  \vspace{-0.15in}
\end{figure}

Vehicle detection is essential in various computer vision applications including traffic surveillance \cite{lu2016optasia} and auto-driving. Classic vehicle detectors \cite{viola2001rapid,felzenszwalb2010object} have achieved promising detection results. Notably, the cascade object detector \cite{viola2001rapid} applies a set of weak-classifiers and filters background objects early. More recently, the region-based RCNN \cite{girshick2014rich} has gained considerable attention in detecting generic objects. Faster RCNN \cite{renNIPS15fasterrcnn} achieves state-of-the-art performance by generating potential object boxes with an embedded region proposal network (RPN). Detecting vehicles robustly and efficiently under different pose, scale, occlusion, and lighting conditions is nevertheless challenging. Figure \ref{fig:gt_face} demonstrates two examples for vehicle detection with real-world traffic camera feeds. To boost the accuracy as well as efficiency for vehicle detection, we are inspired by the classic cascade object detection framework; a novel deep learning framework, namely Evolving Boxes (EB), is proposed in which the object boxes are being refined along with our algorithmic pipeline.

\begin{figure*}[t]
  \centering
  \includegraphics[width=\linewidth]{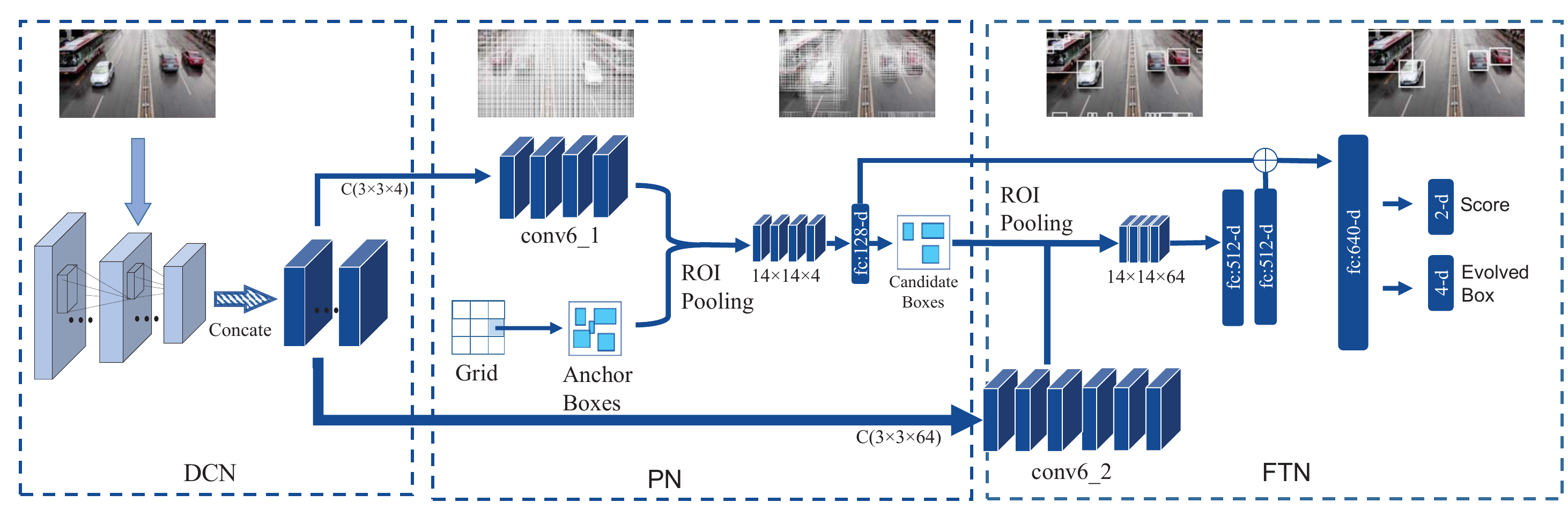}
  \caption{The evolving architecture of our framework. Three networks are involved including the Deep Convolutional Network (DCN), the Proposal Network (PN) and the Fine-Tuning Network (FTN). DCN is responsible for generating rich features from images. PN produces anchor proposals and filters the ones that are unlikely to be vehicles. FTN further fine-tunes the candidates and generates refined localization and recognition results. Concate: feature concatenation from different conv layers.    $\oplus$: feature concatenation from different networks. }
  \vspace{-0.1in}
  \label{fig:architecture}
\end{figure*}

Specifically, we set up two neural networks in a single deep learning framework after a pre-trained deep convolutional network. The first proposal network (PN) employs a set of small convolutional layers to generate candidate boxes while discarding the unlikely ones. Potential boxes are sent to the second fine-turning network (FTN) in which sophisticated convolutional layers are leveraged. Intriguingly, we find that applying different feature fusion techniques yields promising results. That is, by combining the features from PN, FTN and different convolutional layers, the initial object boxes can be refined in terms of both localization and class regression. We demonstrate a considerable improvement for the evolved boxes over those regressed directly from the sophisticated features.

Given such evolving detection structure, we are able to accelerate the overall deep learning networks for vehicle detection. This is especially useful for real-world scenarios where background objects are often the majority. On the other hand, training and using our deep neural networks is end-to-end; it does not require frequently and densely turning the hyper-parameters or in multiple training stages.

To evaluate our proposed framework, we report the evaluations on the recent DETRAC vehicle detection dataset \cite{DETRAC:CoRR:WenDCLCQLYL15}, and we compare with several recent approaches. Notably, we achieve a significant 9.5\% mAP improvement over the state-of-the-art Faster RCNN \cite{renNIPS15fasterrcnn}. Meanwhile, a 9-13 FPS detection rate is obtained on an Nvidia Titan X GPU (Maxwell).

The rest of this paper is organized as follows. Section 2 introduces the background and related work for vehicle detection. Section 3 discusses our evolving framework in detail, while Section 4 demonstrates the experiments.

\section{Related Works}

Vehicle detection has been studied for decades with numerous real applications \cite{song2013discriminatively,feris2015efficient,dollar2014fast}; it has drawn considerable attention when the cascade methods \cite{dollar2014fast} and the deformable part models (DPM) detectors \cite{felzenszwalb2010object} were introduced. The cascaded vehicle detection was initially proposed by Viola and Jones \cite{viola2001rapid} 
with a set of weak classifiers to early filter image patches that are not target objects. Later works \cite{li2015convolutional,qin2016joint,zhang2016joint} extend this cascade pipeline and achieve good performance. On the other hand, DPM detectors successfully detect the target objects by retrieving the object parts and finding their spatial combinations \cite{song2013discriminatively}.
So far, Convolutional Neural Networks (CNNs) have shown their rich representative power in object detection \cite{girshick2014rich}. To employ the cascade detection strategy and to reduce candidate objects, the Faster RCNN \cite{renNIPS15fasterrcnn} achieves state-of-the-art performance by generating potential object locations using RPN. Inspired by these works, we extend the cascade strategy and propose a novel evolving framework for vehicle detection.
Recently, several popular object detection frameworks utilize anchors or grids to propose candidate object locations, which yields end-to-end object detections without explicitly cropping out candidate objects for further recognition. YOLO \cite{redmon2015you} directly regresses the object locations for each grid in the image. SSD \cite{liu2015ssd} generates several candidate objects for each anchor in the image. We find that directly localizing the objects in a single shot often yields unsatisfactory results. Instead, our proposed boxes are evolving and being fine-tuned by different networks within our framework.

\section{Framework}
We illustrate the evolving architecture and different networks of our deep learning framework in Figure \ref{fig:architecture} and as follows:

\begin{itemize}
\item \emph{Deep Convolutional Network} (DCN). The entire images are initially fed to several convolutional and max-pooling layers to generate rich feature representations. Feature concatenation from different convolutional layers are used here. The output feature maps are sent to the proposal network (PN) and the fine-turning network (FTN) afterwards.

\item \emph{Proposal Network} (PN). As shown in Figure \ref{fig:architecture}, the PN network takes the feature maps from the previous convolutional network (DCN) as input. A small convolutional layer with 4 filters later is connected afterwards (conv6\_1). The image and the output feature map are divided into grids; for each grid, a fixed number of anchor proposals are generated. For each anchor proposal, the ROI pooling layer crops out a $14\times14\times4$ feature vector, based on which a fully connected (fc) layer of 128-d directly regresses the initial bounding box $(x, y, w, h)$ and its object score $s$. Candidate boxes with low object scores are discarded.

\item \emph{Fine-tuning Network} (FTN). The FTN network similarly takes the feature maps from the previous convolutional network (DCN) as input. A large convolutional layer with 64 filters is connected afterwards (conv6\_2). For each candidate box produced by the PN network, another ROI pooling layer crops out a $14\times14\times64$ feature vector, which is sent to a set of fc layers for fine-turning. The fc layer feature from the PN network is concatenated to the fc layer feature (512-d) herein. The final output is evolved object boxes $(x', y', w', h')$ and object scores $s'$.

\end{itemize}

We describe below each individual network in detail.

\vspace{0.08in}
\noindent\textbf{Deep Convolutional Network (DCN).} Rich feature representations are produced by this network. We leverage the convolutional and max-pooling layers from the VGG-16 \cite{simonyan2014very} network and load the weights pre-trained on the ImageNet dataset \cite{deng2009imagenet}. This is a common practice among many state-of-the-art deep learning frameworks \cite{redmon2015you,renNIPS15fasterrcnn} to ensure that the detection can benefit from evolving big datasets.

As discussed by \cite{ghodrati2015deepproposal}, high convolutional layers perform better on the classification but lack insight to precise object localization, because the feature weights have been summarized by multiple convolutional layers. On the contrary, low convolutional layers have a better scope to localizing objects as they are closer to raw images. Prior works \cite{ranjan2016hyperface,kong2016hypernet} have shown the gain of combining the feature maps from different convolutional layers. To mirror this, we leverage concatenation of different convolutional layers in DCN to feed into the later PN and FTN networks. Figure \ref{fig:architecture} demonstrates the multi-feature map concatenation (denote as `concate' in Figure \ref{fig:architecture}) and we will show in our experiments how this affects the detection accuracy and speed.  We leverage down- and up-sampling methods when combining different feature maps: we down-sample low layer feature maps to align with high layer feature maps when combining the two layers; to combine three layers, we down-sample the low layer and up-sample the high layer to align with the middle layer feature map. After aligning the feature maps, we normalize these feature maps using a batch normalize layer \cite{ioffe2015batch}, and then concatenate them to form a hyper feature map. In our experiments, we concatenate the feature maps from the 1st, 3rd, and 5th convolutional layers.

\vspace{0.08in}
\noindent\textbf{Proposal Network (PN).}
As background objects are always the majority, filtering out pool candidates is essential to build an efficient vehicle detection system. To this end, we leverage a small PN network to propose candidate boxes as well as to eliminate background regions.
Specifically, a small convolutional layer with 4 filters is connected to the DCN output (conv6\_1) and the $256\times144$ feature map is obtained. We divide the feature maps into $64\times36$ grids while for each grid, a set of \emph{anchor boxes}, or initial object boxes, are generated with fixed sizes of $32\times32$, $64\times64$, $128\times128$, $256\times256$ and $512\times512$, and different aspect ratios of 1:2, 2:1, and 1:1. Unlike the Faster RCNN, we directly take the anchor boxes to propose candidate objects, instead of using an additional fc layer. A ROI pooling layer later crops out a $14\times14\times4$ feature vector from each anchor box. The cropped feature vector is sent to a fc layer with 128-d to regress the candidate object box $(x, y, w, h)$ and object score $s$. We show later in our experiments that using this simple but effective proposal network, roughly 98\% of the background regions can be discarded, leading to a significant speed boost.

\vspace{0.08in}
\noindent\textbf{Fine-tuning Network (FTN)}. The FTN is responsible for fine-tuning the remaining object boxes. The structure is similar to the previous PN network, except that a convolutional layer with 64 filters are used (conv6\_2) to produce the $256\times144$ feature map. The ROI pooling layer extracts a $14\times14\times64$ feature vector for each vehicle candidate produced by the PN. A fc layers with 640-d is connected afterwards that is concatenated from two fc layers (the concatenation is denoted by the $\oplus$ operator in Figure \ref{fig:architecture}); the first has 512-d produced by the current FTN, while the second 128-d is generated by the PN. The output of the fc layers is a 5-d vector, ($x', y', w', h'$) for evolved object box and $s'$ for refined object score. We will show in our experiments that this evolving architecture is superior to directly regressing the object boxes, which is done in many state-of-the-art detection frameworks such as YOLO and Faster RCNN. Further, the concatenation of the two-stage features further improves the detection results.

\vspace{0.08in}
\noindent\textbf{Networks Training}.
As stated above, our networks are partly initialized by the pre-trained ImageNet model VGG-16 \cite{simonyan2014very} for the DCN part, and the other networks are randomly initialized from a zero-mean Gaussian distribution with standard deviation of 0.01. We set the initial learning rate to $10^{-3}$ and then decrease to $10^{-4}$ after 50k iterations. We totally train 70k iterations. The benchmark and comparisons are reported with training using the entire training and validation set, while to evaluate different algorithmic components, only the training set is used.

Our model is trained end-to-end using stochastic gradient descent. We use the mini-batch size of 256. For the PN network, we assign positive anchor proposals that overlap the ground truth for more than 0.5 in intersection over union (IOU) \cite{felzenszwalb2010object}. Anchor proposals that overlap the ground truth for less than 0.3 in IOU are assigned as negative examples. We run non-maximum suppression (NMS) \cite{felzenszwalb2010object} with threshold 0.7 to eliminate redundant boxes and keep 800 of them. For the FTN network, we assign positive candidates that overlap the ground truth for IOU$\ge$0.45, while candidates with $0.1\le$IOU$\le$0.3 are assigned with negative examples. We also apply hard mining \cite{felzenszwalb2010object} during training the FTN; we sort the classification loss in descending order, and pick the top 70\% samples to participate in the back propagation and we ignore easy examples.

\vspace{0.08in}
\noindent\textbf{Multi-stage Loss.}
The fc layers in PN and FTN generate ($x, y, w, h, s$) and ($x', y', w', h', s'$) respectively, denoting the bounding box locations and object scores produced by the two networks. Similar to \cite{girshick2014rich}, we use $t = (t_x, t_y, t_w, t_h)$ to parameterize the bounding box generated by the first stage PN:
\begin{equation}
\begin{split}
  t_x = (x - \hat x) ~/ \hat w, & ~~ t_w = \log (w/\hat w), \\
  t_y = (y - \hat y) ~/ \hat h, & ~~~ t_h = \log (h/\hat h)
\end{split}
\end{equation}
where $\hat x, \hat y, \hat w, \hat h$ are the location of the initial anchor box. Meanwhile, we use $t' = (t'_x, t'_y, t'_w, t'_h)$ to parameterize the evolved box generated by the second stage FTN:
\begin{equation}
\begin{split}
  t'_x = (x' - x) ~/ w, & ~~ t'_w = \log (w'/w),  \\
  t'_y = (y' - y) ~/ h, & ~~~ t'_h = \log (h'/h)
\end{split}
\end{equation}

Therefore, a multi-stage loss $L$ is used to jointly train the two-stage classification and regression:
\begin{equation}
\begin{split}
L & = \alpha L_{pn} + (1 - \alpha)L_{ftn}, \\
L_{pn}(t, s) & = L_{cls}(s) + \lambda L_{loc}(t, s), \\
L_{ftn}(t', s') & = L_{cls}(s') + \lambda L_{loc}(t', s'), \\
\end{split}
\label{affineTransform}
\end{equation}
where $L_{pn}$ and $L_{ftn}$ are the loss for PN an FTN respectively. ${\alpha}$ balances the two stages. The class regression loss $L_{cls}(s) = -\log s$ is a logarithmic loss upon class scores. To regress bounding boxes locations, we follow \cite{girshickICCV15fastrcnn} and use the localization loss $L_{loc}$ defined as:
\begin{equation}
  L_{loc}(t, s) = \sum_{i \in \{x, y, w, h\}} \sigma(t_i - s),
\end{equation}
where
\begin{equation}
\sigma(x)=\left\{
\begin{array}{ll}
0.5x^2 & \text{if~} |x| < 1,\\
|x| - 0.5 & \text{otherwise.}
\end{array} \right.
\end{equation}

We consider the PN and FTN stages equally important; $\alpha = 0.5$ is set to compute the multi-stage loss.

\section{Experiments}
\label{sec:exp}
We trained our model on the recent DETRAC vehicle detection dataset \cite{DETRAC:CoRR:WenDCLCQLYL15} with 140K captured frames and 1.2M labeled vehicles. It contains 84K images for training and we further split it into the training set with 56K images and the validation set with 28K image. The image resolution is $960\times540$, and user marked rectangles exist to represent non-detection regions. The DETRAC dataset is challenging due to its large variation; the cameras are mounted on traffic poles in Beijing, while the video frames are captured in different scenarios including sunny, cloudy, rainy and night. By mean each video frame contains 8.6 vehicles and occlusion happens frequently. Figure \ref{fig:gt_face} shows two examples of this dataset.

To evaluate the effectiveness of our proposed framework, we focus on answering (1) how different algorithmic components affects the vehicle detection performance, and (2) how our overall framework compares with state-of-the-art vehicle detection methods.  For a fair comparison, results for the first experiment is reported on the validation set while benchmarks on the test set is reported to compare with other methods. The following sections discuss each experiment respectively.

\subsection{Control Experiments}
Table \ref{tb:control} demonstrates the control experiments of switching off different components of our proposed framework. Detection performances including the overall mAP and mAPs under different scenarios are reported. PN+FTN+Fusion indicates that we do not use the multi-layer feature map concatenation from the 1st, 3rd, and the 5th convolutional layers and only the final convolutional layer features are used. PN+FTN+Concat turns off the multi-stage feature concatenation from different networks, while our full model is PN+FTN+Fusion+Concat. We also compare with the Faster RCNN \cite{renNIPS15fasterrcnn}, which can be seen as using the PN to propose candidate vehicles but directly regressing the vehicle detection results; fine-tuning the detections boxes and class scores are not involved in their approach. Besides, multi-layer and multi-stage feature concatenation are not used in Faster RCNN either.

\begin{table}[t]
\footnotesize
\begin{center}
\begin{tabular}{c|c|P{0.2in}P{0.2in}P{0.2in}P{0.2in}}
\hline
Algorithmic Setting & Overall & Sunny & Cloudy & Rainy & Night \\
\hline
Faster RCNN \cite{renNIPS15fasterrcnn} & 68.58 & 63.64 & 70.04 & 81.56 & 60.53\\
PN+FTN+Fusion & 73.96 & 69.70 & 73.90 & 82.12 & 67.91\\
PN+FTN+Concat & 83.84 & 87.09 & 84.80 & 85.95 & 70.21\\
PN+FTN+Fusion+Concat & \textbf{84.43} & \textbf{87.48} & \textbf{85.88} & \textbf{85.65} & \textbf{70.86}\\
\hline
\end{tabular}
\end{center}
\vspace{-0.1in}
\caption{Control experiments on switching off different algorithmic components of our framework. We illustrate mean average precisions (mAP) on the DETRAC validation set as well as different subsets. }
\vspace{-0.1in}
\label{tb:control}
\end{table}

\begin{table}[t]
\begin{center}
\begin{tabular}{c|c}
\hline
Algorithmic Setting & Runtime Speed (ms) \\
\hline
Faster RCNN \cite{renNIPS15fasterrcnn} & 87\\
PN+FTN+Fusion & 75\\
PN+FTN+Concat & 100\\
PN+FTN+Fusion+Concat & 110\\
\hline
\end{tabular}
\end{center}
\vspace{-0.1in}
\caption{Inference time on different algorithmic combinations. }
\label{tb:speed}
\vspace{-0.1in}
\end{table}

\begin{table*}[t]
\small
\begin{center}
\begin{tabular}{c|c|P{0.35in}P{0.35in}P{0.35in}|P{0.35in}P{0.35in}P{0.35in}P{0.35in}|P{0.5in}P{0.85in}}
\hline
Method & Overall & Easy & Medium & Hard & Sunny & Cloudy & Rainy & Night & Speed & Environment\\
\hline
DPM \cite{felzenszwalb2010object}& 25.70 & 34.42 & 30.29 & 17.62 & 24.78 & 30.91 & 25.55 & 31.77 & 6s/img & CPU@2.4GHz \\
ACF \cite{dollar2014fast}& 46.35 & 54.27 & 51.52 & 38.07 & 58.30 & 35.29 & 37.09 & 66.58 & 1.5s/img & CPU@2.4GHz\\
RCNN \cite{girshick2014rich}& 48.95 & 59.31 & 54.06 & 39.47 & 59.73 & 39.32 & 39.06 & 67.52 & 10s/img & GPU@K40\\
Faster RCNN \cite{renNIPS15fasterrcnn} & 58.45 & 82.75 & 63.05 & 44.25 & 62.34 & 66.29 & 45.16 & 69.85 & 0.09s/img & GPU@TitanX\\
CompACT \cite{cai2015learning}& 53.23 & 64.84 & 58.70 & 43.16 & 63.23 & 46.37 & 44.21 & 71.16 & 4.5s/img & GPU@K40\\
Ours - EB& \textbf{67.96} & \textbf{89.65} & \textbf{73.12} & \textbf{54.64} & \textbf{72.42} & \textbf{73.93} & \textbf{53.40} & \textbf{83.73} & 0.11s/img & GPU@TitanX\\
\hline
\end{tabular}
\end{center}
\vspace{-0.1in}
\caption{Mean average precision (mAP) on the DETRAC test dataset produced by different state-of-the-art vehicle detection approaches. The runtime environment and speed are shown as well. }
\label{tb:all}
\vspace{-0.1in}
\end{table*}
Quantitatively speaking, our full model performs the best among different algorithmic settings. As the Faster RCNN does not apply the proposal refinement or any other tricks discussed in this paper, its performance drops significantly for a 15.9\% mAP compared with our full model. The multi-layer feature map fusion is critical, which introduces a 10.5\% performance gain. This accords with the study in \cite{ghodrati2015deepproposal}, and we consider this particularly useful for vehicle detection, as many vehicles from real traffic cameras are small in size. Ignoring the low level convolutional layer features prevents the network from finding any small object.  Further, the multi-stage feature concatenation leads to another 0.6\% performance gain. On the other hand, although feature fusion introduces additional runtime, the overall speed is still acceptable. A 9-13 FPS detection rate is achieved using a single Nvidia Titan X GPU (Maxwell).

\subsection{Comparing with State-of-The-Art}

\begin{figure}[t]
  \centering
  \includegraphics[width=0.95\linewidth]{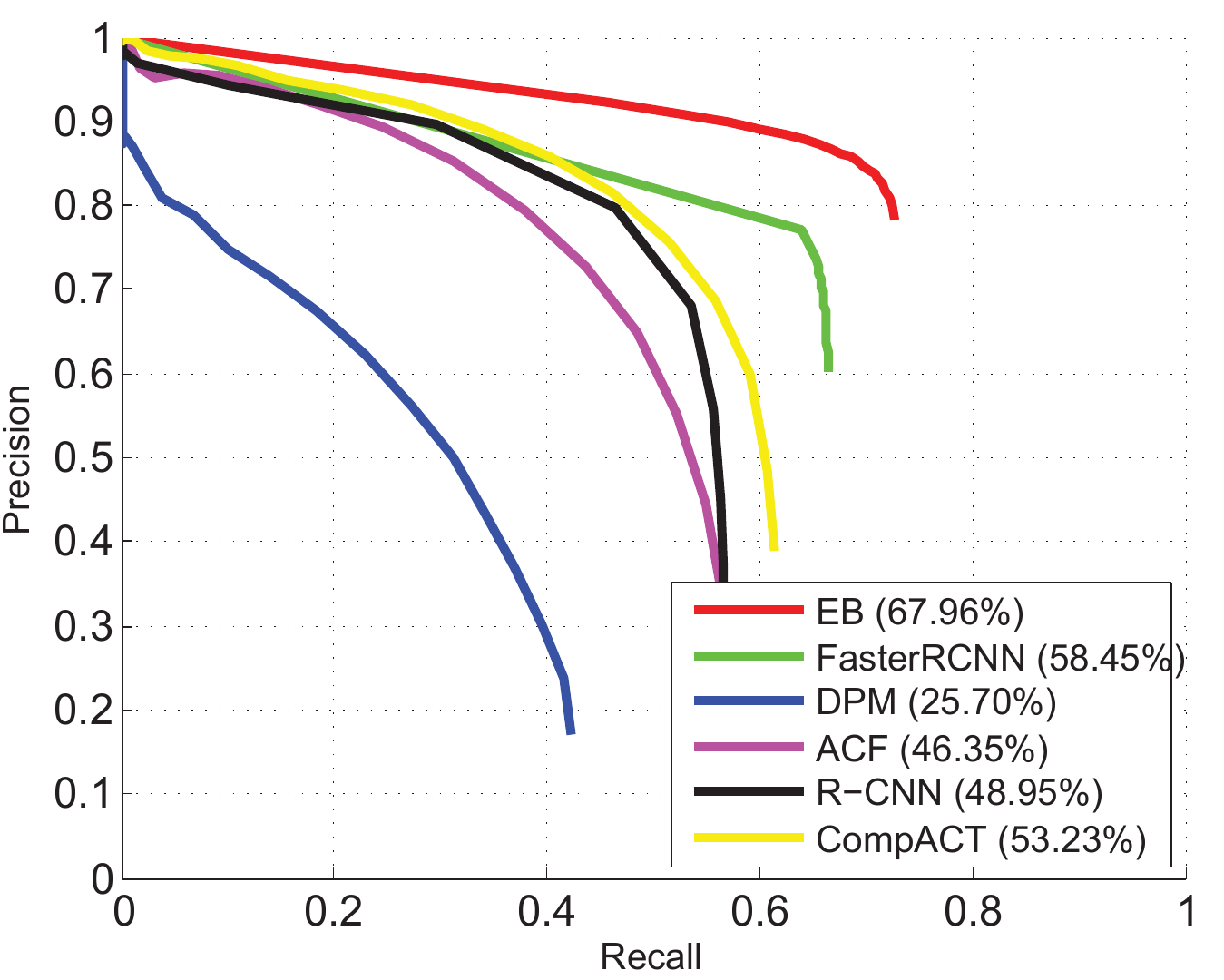}
  \caption{Precision-recall curves of different vehicle detection algorithms on the DETRAC test set.}
  \label{fig:pr}
  \vspace{-0.1in}
\end{figure}

Figure \ref{fig:pr} and Table \ref{tb:all} demonstrate the comparison of our full model with state-of-the-art vehicle detection approaches. The results can also be found at the DETRAC benchmark server\footnote{\url{http://detrac-db.rit.albany.edu/DetRet}}. Precision-recall curves and mAPs are reported. We compare with CompACT \cite{cai2015learning}, RCNN \cite{girshick2014rich}, ACF \cite{dollar2014fast}, Faster RCNN \cite{renNIPS15fasterrcnn}, and DPM \cite{felzenszwalb2010object}. We achieve a significant overall improvement of 14.73\% mAP over the state-of-the-art CompACT \cite{cai2015learning} and 9.5\% mAP over Faster RCNN \cite{renNIPS15fasterrcnn}. Notably, our method performs the best on all subcategories. Figure \ref{fig:pr} further shows that our methods outperforms state-of-the-art approaches with different recall setting, indicating that our method achieves better detection coverage as well as accuracy.

Table \ref{tb:all} shows the environment and runtime speed for different approaches. Our proposed framework runs magnitude faster compared with CompACT (40x) and RCNN (90x), while on-par or slightly slower than the Faster RCNN.

\begin{figure*}[t]
  \centering
  \includegraphics[width=0.19\linewidth]{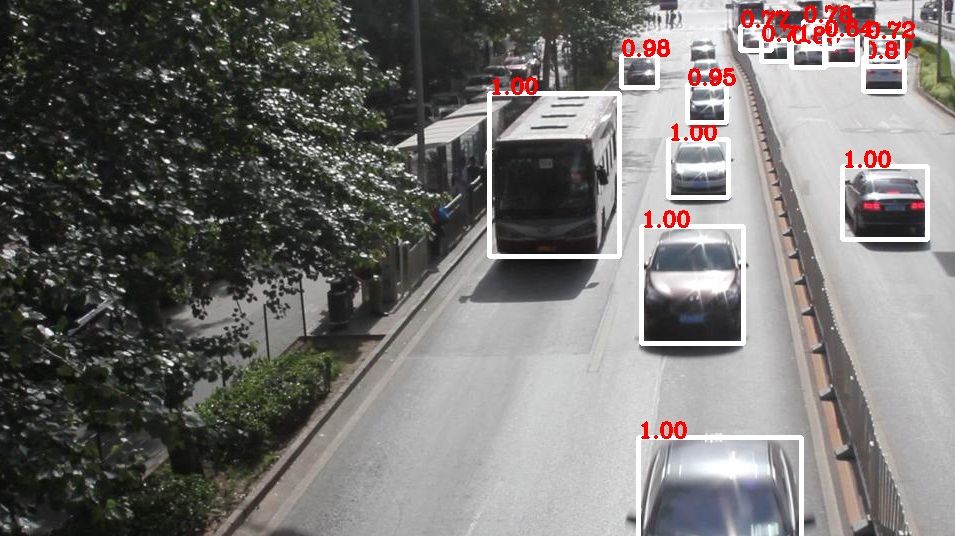}
  \includegraphics[width=0.19\linewidth]{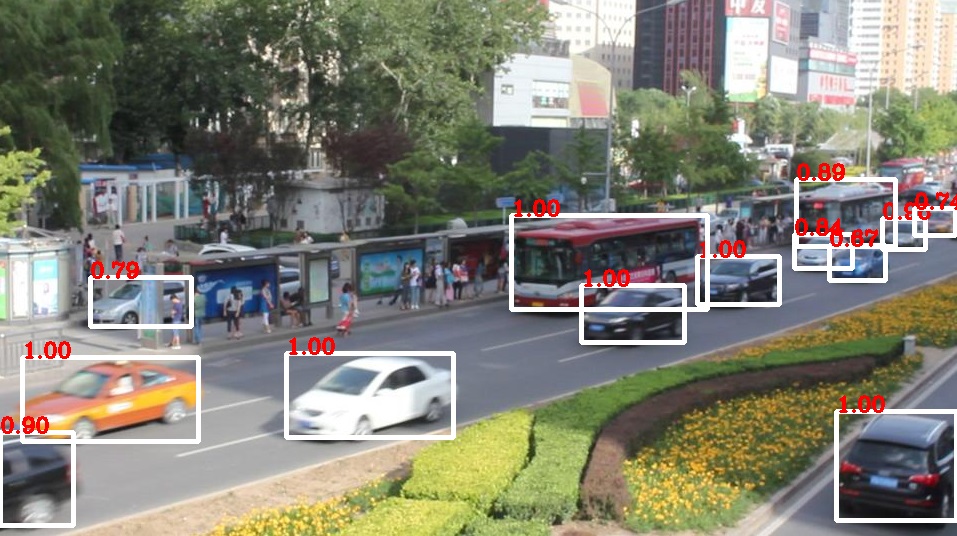}
  \includegraphics[width=0.19\linewidth]{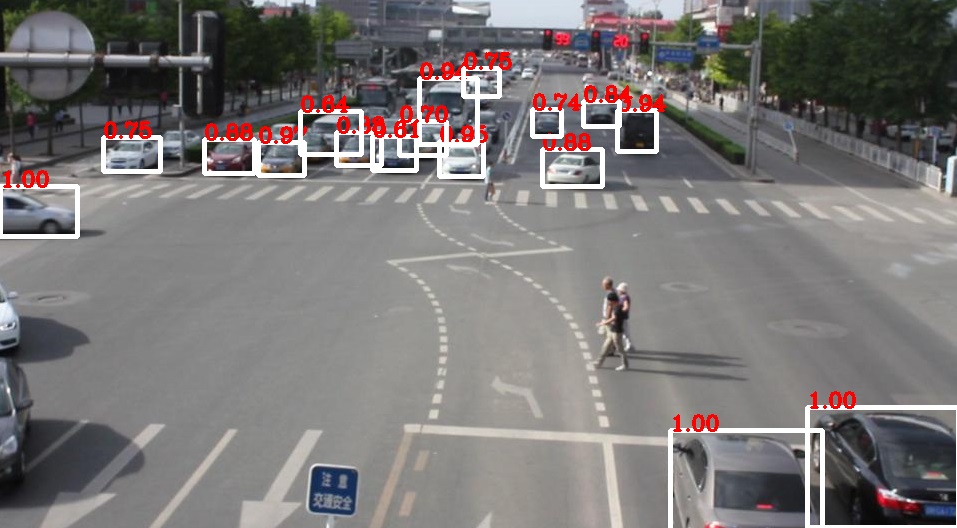}
  \includegraphics[width=0.19\linewidth]{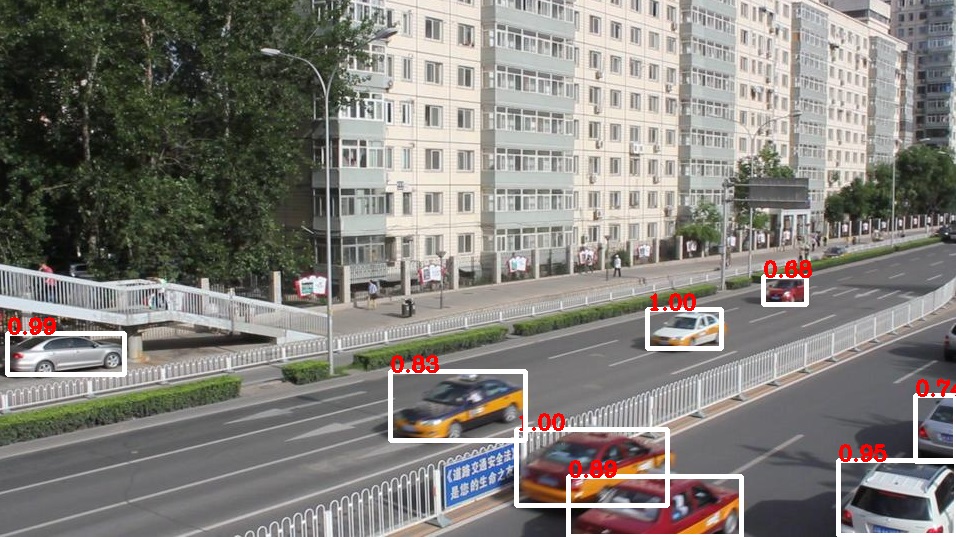}
  \includegraphics[width=0.19\linewidth]{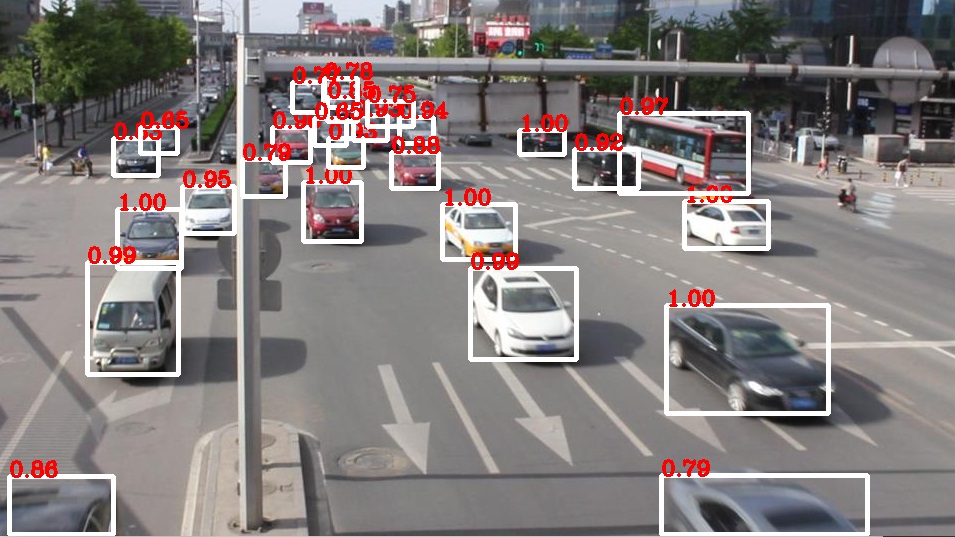}

  \includegraphics[width=0.19\linewidth]{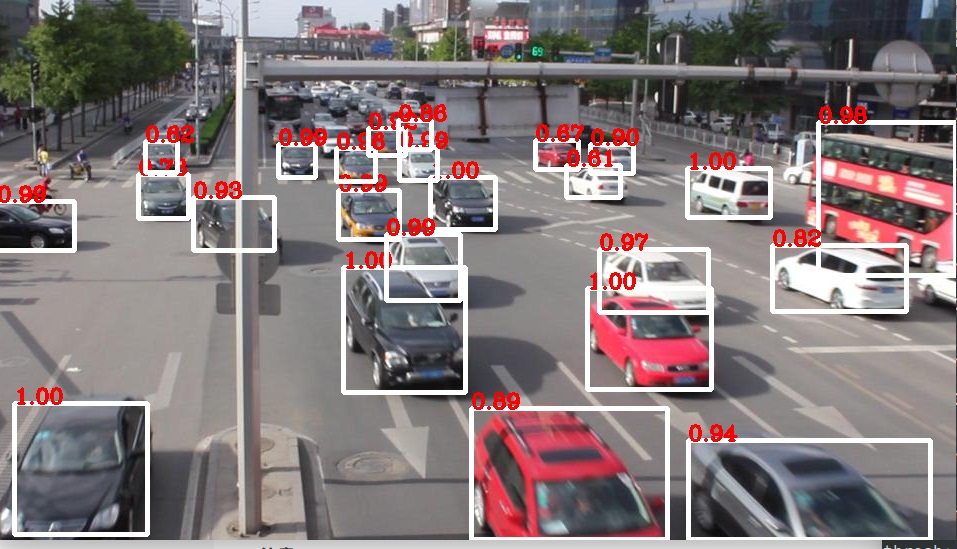}
  \includegraphics[width=0.19\linewidth]{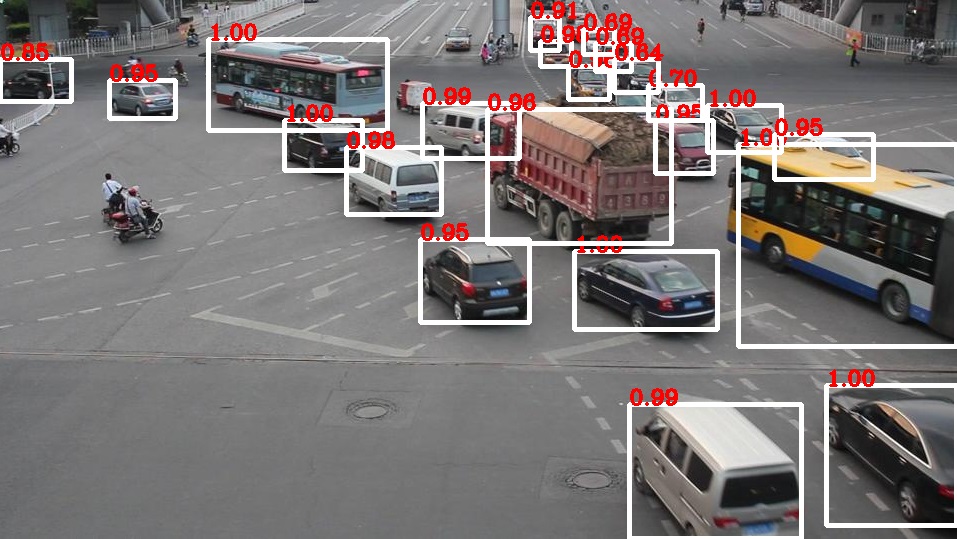}
  \includegraphics[width=0.19\linewidth]{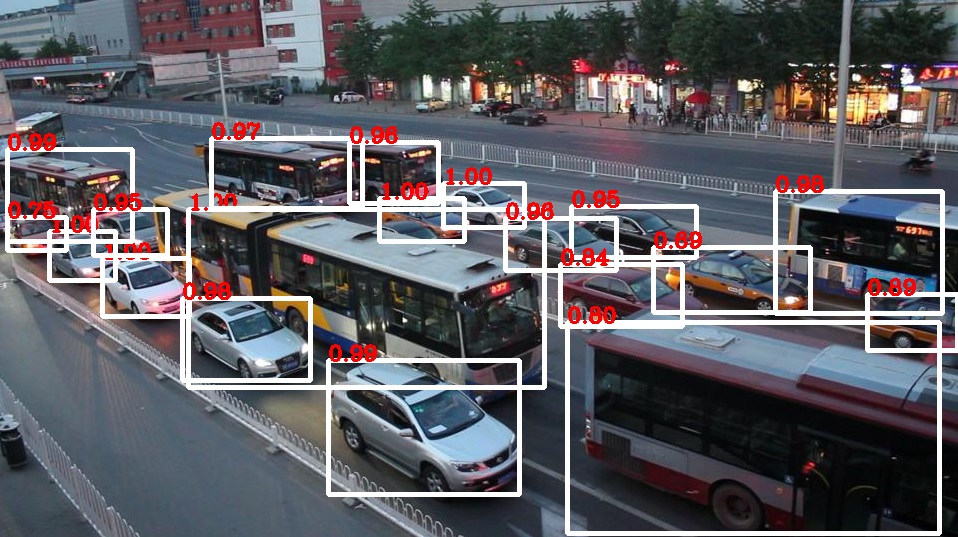}
  \includegraphics[width=0.19\linewidth]{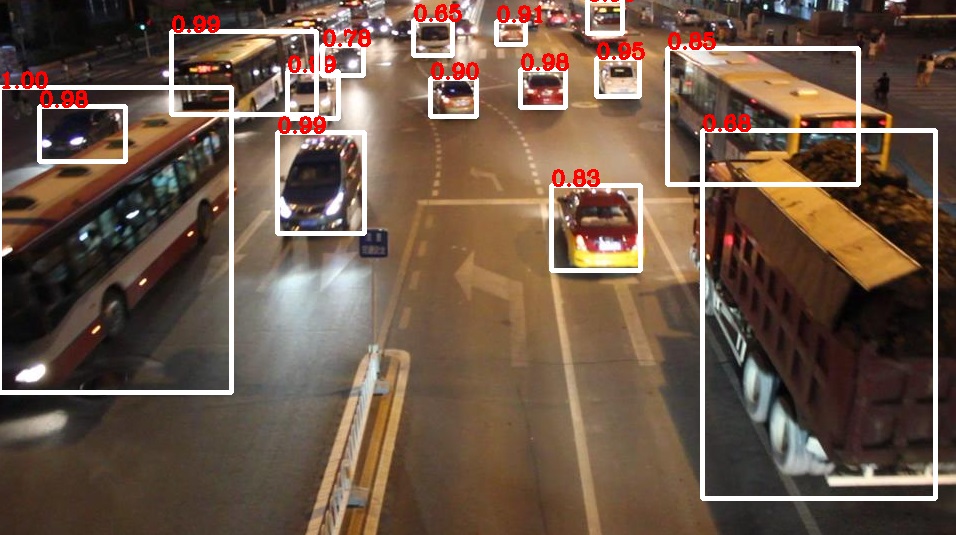}
  \includegraphics[width=0.19\linewidth]{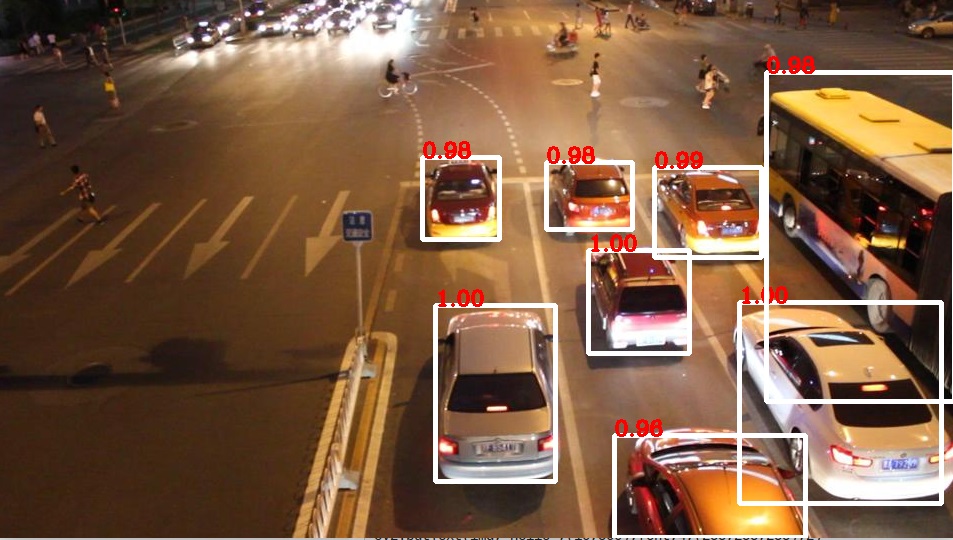}
  \caption{Successful detection results on the DETRAC test set. }
  \label{fig:good}
\end{figure*}

\begin{figure*}[t]
  \centering
  \includegraphics[width=0.19\linewidth]{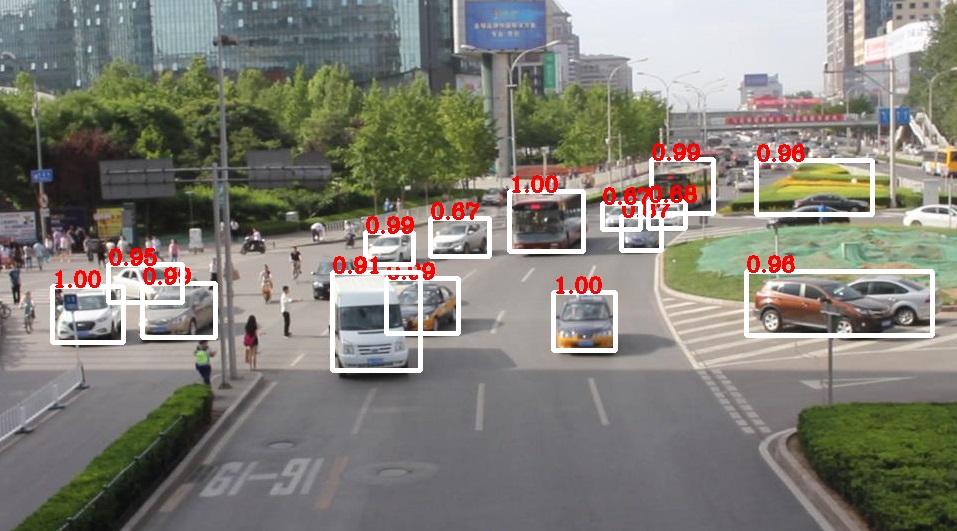}
  \includegraphics[width=0.19\linewidth]{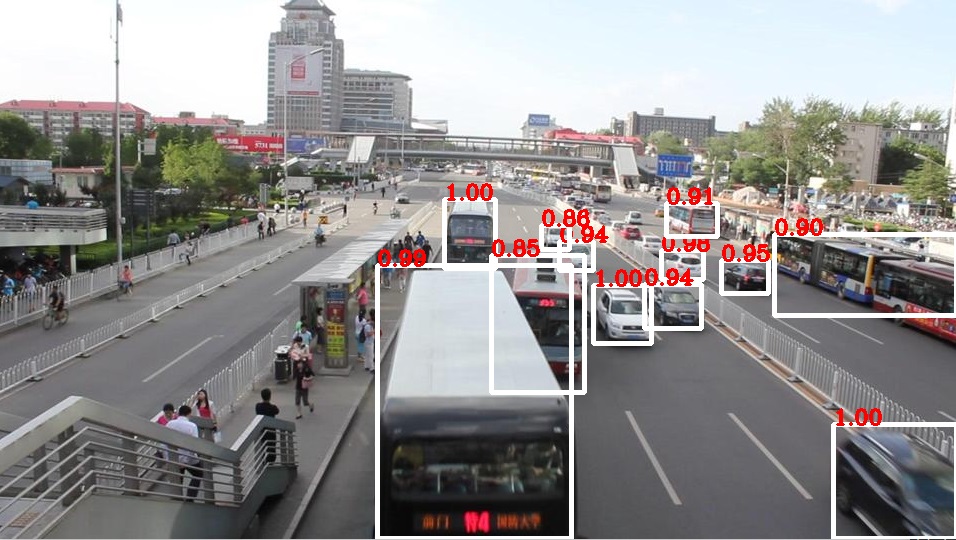}
  \includegraphics[width=0.19\linewidth]{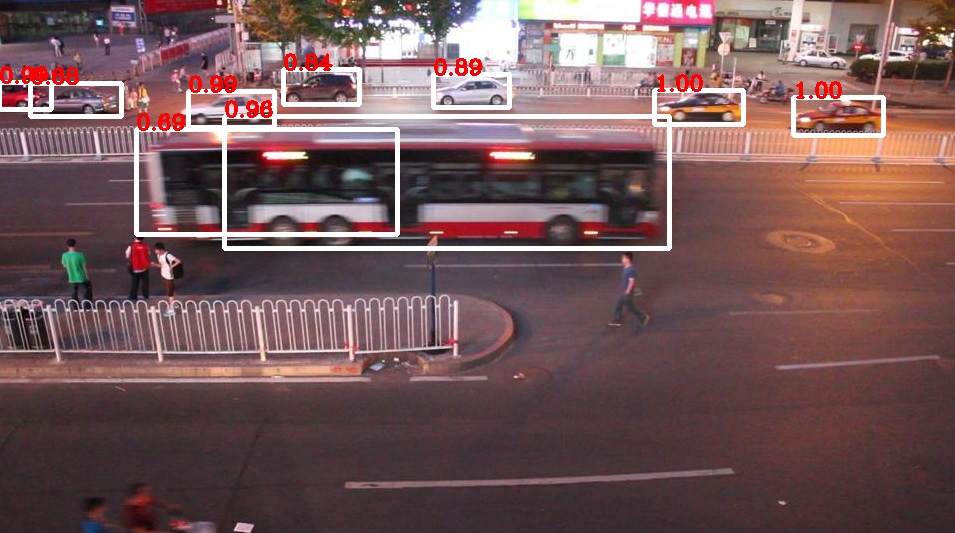}
  \includegraphics[width=0.19\linewidth]{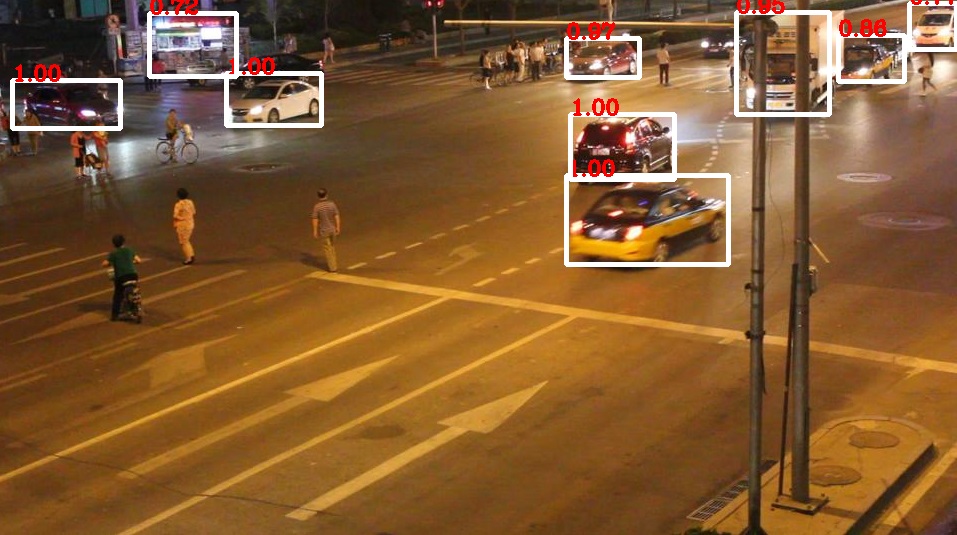}
  \includegraphics[width=0.19\linewidth]{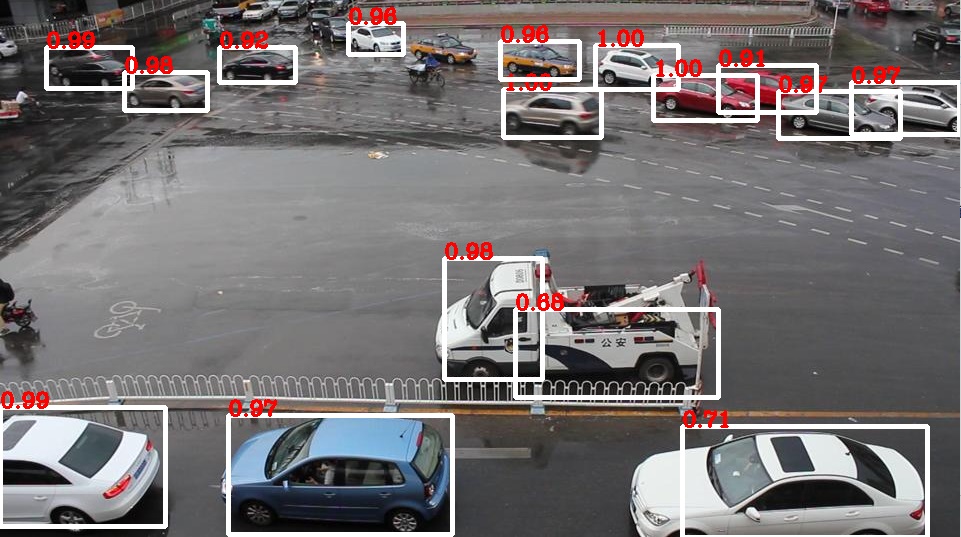}
  \caption{Partially unsuccessful detection results on the DETRAC test set.}
  \label{fig:bad}
\end{figure*}

Figure \ref{fig:good} and \ref{fig:bad} demonstrate qualitative evaluations of our approach on the test set; successful and partially unsuccessful results are shown. We succeed in detecting most of the vehicles in different appearances, especially when heavy occlusion is happening or the vehicles are far away from the camera. However, there are also some failure cases where the vehicle detection is split into multiple boxes, or fails to identify multiple vehicles that are adjacent to each other. Generally speaking, the detection results are reasonable and high quality for further post-processing such as vehicle type and color recognition.

Our framework is implemented on Caffe \cite{jia2014caffe}, and we have released the code and trained model for future research\footnote{\url{http://zyb.im/research/EB}}.

\section{Conclusions}
  We borrow the idea from cascade object detection and propose an evolving object detection framework in which the object boxes are generated and being refined by different networks within our proposed pipeline. We show that by leverage different feature fusing techniques, good performance is achieved for both localization and class recognition. The runtime speed is 9-13 FPS on a moderate commercial GPU.

\balance
\bibliographystyle{IEEEbib}
\bibliography{EB_arXiv}

\end{document}